\documentclass{article}
\usepackage{spconf,amsmath,graphicx,amsfonts}
\usepackage{multirow}
\usepackage{booktabs}
\usepackage{multirow}

\title{EmoTalker: Emotionally Editable Talking Face Generation via Diffusion Model}
\name{Bingyuan Zhang$^{1,2\ddagger}$\thanks{$^\ddagger$ Both authors have equal contributions. }, Xulong Zhang$^{1\ddagger}$, Ning Cheng$^{1\ast}$\thanks{$^\ast$ Corresponding authors: Ning Cheng (chengning211@pingan.com.cn), Jun Yu (harryjun@ustc.edu.cn)}, Jun Yu$^{2\ast}$, Jing Xiao$^{1}$, Jianzong Wang$^{1}$}

\address{$^{1}$Ping An Technology (Shenzhen) Co., Ltd., Shenzhen, China\\
$^{2}$University of Science and Technology of China, Hefei, China}
\begin{document}
%
\maketitle
\vspace{-20pt} 
\begin{abstract}
In recent years, the field of talking faces generation has attracted considerable attention, with certain methods adept at generating virtual faces that convincingly imitate human expressions. 
However, existing methods face challenges related to limited generalization, particularly when dealing with challenging identities. 
Furthermore, methods for editing expressions are often confined to a singular emotion, failing to adapt to intricate emotions.
To overcome these challenges, this paper proposes EmoTalker, an emotionally editable portraits animation approach based on the diffusion model. 
EmoTalker modifies the denoising process to ensure preservation of the original portrait's identity during inference. 
To enhance emotion comprehension from text input, Emotion Intensity Block is introduced to analyze fine-grained emotions and strengths derived from prompts. 
Additionally, a crafted dataset is harnessed to enhance emotion comprehension within prompts.
Experiments show the effectiveness of EmoTalker in generating high-quality, emotionally customizable facial expressions.
\end{abstract}
\vspace{-5pt} 

\vspace{-10pt} 
\section{Introduction}
\label{sec:intro}
\vspace{-10pt} 
Talking face generation \cite{prajwal2020lip, zhou2020makelttalk, zhang2022Shallow, wang2023CP-EB,sun2022Pre-Avatar} uses deep learning to create realistic faces that mimic human faces, resulting in engaging virtual assistants and chatbots. 
This technology has a wide range of applications and can greatly improve the user experience.


Numerous researchers tend to neglect the undertaking of intricate emotion-driven talking face generation, a critical aspect in crafting animated faces that are not only realistic but also capable of conveying a wide range of emotions with remarkable expressiveness.
While some talking face generation research \cite{vougioukas2020realistic, chen2020talking, Ji_2021_CVPR}  focuses on incorporating facial emotions, it may not always provide the ability to edit or customize these emotions freely. 
EVP \cite{Ji_2021_CVPR} not only derives emotions from speech through disentanglement but also explores the approach of emotion editing via interpolation within the emotional feature space. 
This approach offers valuable insights into the field of emotion manipulation.
However, other advancements \cite{eskimez2021speech, wang2020mead, sinhaemotion, Xu_2023_CVPR, papantoniou2022neural} in the field have successfully integrated editing emotions into talking face generation, some \cite{ji2022eamm, liang2022expressive} rely on a referred image to capture emotions. 
In the past, initial efforts \cite{Ji_2021_CVPR, wang2020mead} in the field of emotional talking face generation were customized to specific identities, while recent developments \cite{ji2022eamm, liang2022expressive} have shifted their focus towards one-shot setting and drive facial expressions by reference faces.

\begin{figure*}[t]
    \centering
    \includegraphics[scale = 0.3]{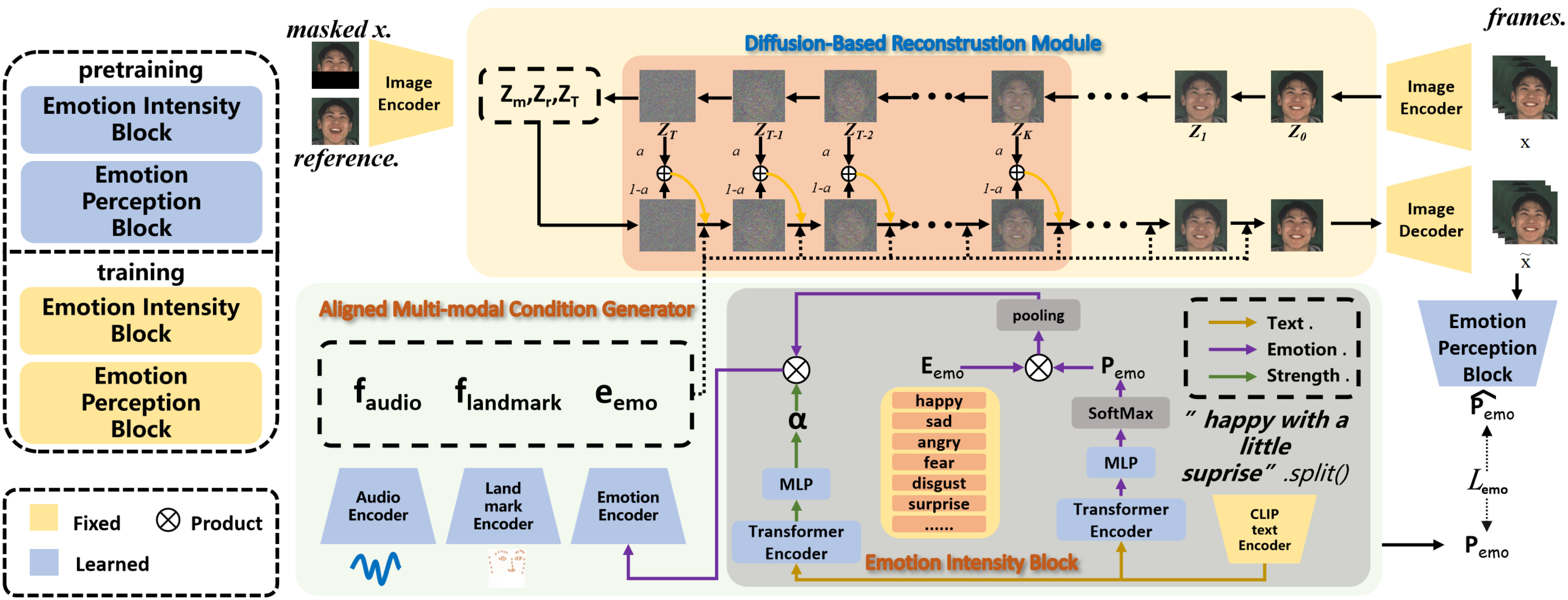}
    \caption{The overall framework of EmoTalker. The condition generator produces conditions to guide the denoising process. And we make the emotions of the generated images approach to those of the prompts.}
    \vspace{-10pt} 
    \label{framework}
\end{figure*}

In the field of talking face generation, earlier research \cite{shen2023difftalk} has persistently grappled with the challenge of inadequate generalization capabilities when dealing with intricate or demanding identities. 
Furthermore, the previous multi-modal emotion editing methodologies \cite{Xu_2023_CVPR, xu2023multimodaldriven} employed in talking face generation were restricted to altering a solitary strength of a particular emotion, resulting in a limitation in the scope of emotion manipulation.

To solve the aforementioned challenges, this paper proposed EmoTalker, building upon the foundation established by Difftalk \cite{shen2023difftalk}.
EmoTalker modifies the denoising mechanism in diffusion models to preserve the original portrait's identity. 
Otherwise, EmoTalker has devised an approach to translate prompts into corresponding emotions and strength values.
Furthermore, we have curated a dedicated dataset to facilitate these advancements.

Our main contributions are as follows:

\vspace{-8pt} 
\begin{itemize}
\setlength{\itemsep}{0pt}
    \item We propose a specially crafted conditional diffusion model, aiming to direct the denoising process towards the desired facial expressions using textual prompts containing intricate emotions and strengths.
    \item To address the challenge posed by the limited generalization ability of challenging portraits, we have modified the denoising process during inference to generate frames that closely align with the portrait's inherent identity.
    \item To facilitate the model's comprehension of intricate emotions and strengths within prompts, we propose Emotion Intensity Block and a new dataset FED.
    The FED dataset is accessible through the following link: https://largeaudiomodel.com/fed/.
\end{itemize}

\section{Method}
\vspace{-5pt} 
The overall framework is shown in Figure~\ref{framework}, wherein the Aligned Multimodal Condition Generator leverage information from diverse modalities, including audio, images, and text, to generate condition. 
This generated condition then guides the Diffusion-Based Reconstruction Module in generating the desired facial expressions.

\vspace{-10pt} 
\subsection{Diffusion-based Reconstruction Module}
Derived from DDIM(Denoising diffusion implicit models), Diffusion-Based Reconstruction Module builds upon the process of noising and denoising within latent spaces through tamming \cite{esser2020taming}. 
It harnesses conditions to guide the animation of portraits.
The image encoder and decoder are both frozen and well-trained, which is designed in alignment with \cite{esser2020taming}.
The image encoder translates the image into a latent space, enabling both noising and denoising processes to occur within compact latent spaces. 
This reduction in computational demand accelerates the model's processing speed.
The landmark encoder $E_L$ is MLP based.
The audio feature obtaining followed DFRF \cite{shen2022learning}, and audio feature was laterly encoded by audio encoder $E_A$.

In EmoTalker, Diffusion-Based Reconstruction Module is harnessed for noise prediction, with the corresponding loss formulated as follows:

\vspace{-10pt} 
\begin{equation}
\begin{aligned}
\mathcal{L}_{EmoDiff}:=\mathbb{E}_{z, \epsilon \sim \mathcal{N}(0,1), C, t}\left[\left\|\epsilon-\mathcal{M}\left(z_t, t, C\right)\right\|_2^2\right],
\end{aligned}
\label{emotalkerloss}
\end{equation}
where $C$ is the condition set of EmoTalker to be explained in Sec. 2.2, $\epsilon$ denotes noise to predict, the $\mathcal{M}$ represents a time-conditional UNet-based denoising network, $t$ and $z_t$ signifies time step and image that is noised at step $t$, and the the network parameters of $\mathcal{M}$, $E_L$, $E_A$ and emotion encoder $E_E$ are optimized via Eq. \ref{emotalkerloss}.

\subsection{Aligned Multi-modal Condition Generator}
\textbf{Condition of Cross-Attention.}
During the denoising process, we leverage audio features $f_{audio}$, landmark features $f_{landmark}$, and emotion embedding $e_{emo}$ as the condition to guide the denoising procedure.
We have reorganized the cross-attention condition, as the key and value of cross-attention, as the Eq. \ref{concat}.

\begin{equation}
\begin{aligned}
C=[f_{audio}, f_{landmark}, e_{emo}],
\end{aligned}
\label{concat}
\end{equation}

By incorporating a strength coefficient multiplied by the condition within the denoising cross-attention mechanism, it becomes feasible to regulate the intensity of the condition's impact \cite{hertz2022prompt}. 
We integrate this mechanism into EmoTalker to exert control over the emotional strength.

\textbf{Emotion Intensity Block.} 
As the text encoder in CLIP \cite{radford2021learning} has a comprehensive semantic space, we employ the frozen CLIP text encoder to serve as our prompt encoder.
Nonetheless, as selecting the one with the highest score in a sequence may result in the loss of certain intricate details, which the CLIP text encoder did, we opt to split each sentence before putting into the prompt encoder.


Upon feeding the prompt into the Emotion Intensity Block, the emotion strength $\alpha$ is acquired via the utilization of the pretrained strength predictor.
The prompt embedding is similarly fed into an emotion predictor to make predictions about emotions. 
The resulting output provides the likelihood for each emotion category.

From the predicted emotion categories, a single emotion is chosen, referred to as the hard label. 
Nonetheless, classifying emotions based on text input presents challenges due to the nuanced relationship between text and emotions. 
Because, a single sentence could convey varying emotions with differing intensities, thereby rendering the application of hard labeled sentiment classifiers incapable of precisely capturing the nuanced emotional nuances within the text.
To address this issue, we employ the utilization of soft label.

Upon acquiring the probabilities for each emotion, we multiply each emotion's likelihood by its corresponding emotion embedding, acquired by the prompt encoder. 
Subsequently, a pooling operation is executed to derive a comprehensive representation of the potential emotional state.
Afterward, we multiply the emotion strength with the comprehensive representation mentioned earlier \cite{hertz2022prompt}, as followed in Eq. \ref{optim4}. 
This process yields the ultimate condition about emotion.

\vspace{-10pt} 
\begin{equation}
\begin{aligned}
e_{\text {emo }}=\sum_i^M\left(P_{\text{emo}, i} \cdot e_i\right)\cdot \alpha,
\end{aligned}
\label{optim4}
\end{equation}
where $P_{\text{emo}, i}$ is the possibility from the softmax layer of the emotion predictor for predicting the input text to every base emotion, $e_i \in E_{emo}$ is the embedding corresponding to every single emotion, $\alpha$ is the strength, $M$ is the number of emotions, and $e_{\text {emo }}$ is the output of Emotion Intensity Block as the final emotion condition.

\textbf{Emotion Perception.}
To produce intricate emotional facial expressions that align with the given prompt, we incorporated an emotion perception block into our model. 
This block is a frozen EfficientNet pretrained by expressive face images.

We compute the cross entropy loss between the Emotion Perception Block's output and the label of the emotion to optimize the overall network. 
The label of the emotion could be obtained by the emotion predictor, if we have proper datasets containing prompts with corresponding videos.
However, because we only have datasets containing specific emotions with corresponding videos, the label of the emotion is only a hard label during training.

\begin{equation}
\begin{aligned}
\mathcal{L}_{\text{emo}} = -\sum_{i}^M P_{\text{emo}, i} \log(\hat{P}_{\text{emo}, i}),
\end{aligned}
\label{entropyloss}
\end{equation}
where $\hat{P}_{\text{emo}, i}$ is the possibility from the Emotion Perception Block, ${P}_{\text{emo}, i}$ represents the label of the emotion, and $i$ denotes the index of emotion of base emotions.

\vspace{-10pt} 
\subsection{Loss Function and Inference}

\vspace{-5pt} 
\textbf{Traning Loss.} 
The loss function employed during EmoTalker's training is as follows:
\begin{equation}
\begin{aligned}
\mathcal{L} = \mathcal{L}_{EmoDiff} + \mathcal{L}_{\text{emo}},
\end{aligned}
\label{entropyloss}
\end{equation}

During the training process, the pretrained Emotion Intensity Block doesn't engage in training, but engages in inference. 
The embeddings of emotion are input into the emotion encoder $E_E$, which is done with the aim of reducing the computational complexity of the model.

\textbf{Expression Generation Preserving Facial Information during Inference.}
The Diffusion model has demonstrated impressive capabilities in generating high-quality images. 
Notably, the unique denoising process of the Diffusion model involves losing some local image details while retaining outline information, presenting an interesting opportunity for facial emotion expression editing.

When the image is noised to a point where it loses information as an almost noise, we designate the noise as $Z_{T}$. 
The $Z_{T}$ image serves as the foundation for denoising. 
However, during denoising, there is a potential risk of sacrificing the original identity information. 
To counter this, we introduce the method of preserving facial information during inference, denoted as adding $\hat{z_t}$, which is the prediction at the step $t$, with $z_t$ in a specific ratio $a$,  as illustrated in Figure~\ref{framework} and Eq. \ref{denoising}. This approach aids in preserving the core semantic details of the original image during inference.

\vspace{-10pt} 
\begin{equation}
\begin{aligned}
\hat{z_{t-1}}=Denoising(z_t*a+\hat{z_t}*(1-a)),
\end{aligned}
\label{denoising}
\end{equation}

The method of preserving facial information is executed from $Z_{T}$ to $Z_{K}$.
Subsequent to $Z_{K}$, the denoising process adheres to the conventional procedure.
Empirically, We set $T$ to 1000, $K$ to 0.65$T$, and preserving ratio $a$ to 0.3.

\vspace{-10pt} 
\section{Experiment}
\vspace{-5pt} 
\subsection{Experiment Dataset and Evaluation Metrics}

\textbf{Dataset.}
We have selected 24 subjects of diverse ethnicity from MEAD \cite{wang2020mead} at level 3 for training 25 epochs.
And we evaluate on both MEAD and CREMA-D \cite{cao2014crema}.

For training the Emotion Intensity Block, we employed the Facial Emotion Description (FED) dataset that we proposed. 
This dataset comprises rows of sentences describing emotions along with their corresponding emotional values. 
Assigning emotional values to sentences is manually determined based on adverbs, adjectives, nouns, and events, as determined by human judgment and linguistic analysis.

\textbf{Metrics.}
We have selected identity CSIMD ($\downarrow$) and Emotion Accuracy ($\uparrow$) as our metrics.
We compute CSIMD(cosine similarity distance) between ArcFace features \cite{deng2019arcface} of the predicted frame and the input identity face of the target.
The emotion classifier network from EVP \cite{Ji_2021_CVPR} is employed to quantify the accuracy of the emotions generated in the final animation. 

\vspace{-10pt} 
\subsection{Comparison with Methods of the SOTA}
We present the quantitative comparison in Table \ref{tab:metrics_comparison}.
In terms of identity CSIMD($\downarrow$), our method is better than MEAD, benefiting from preserving facial information during the inference.
But EVP surpasses our approach in CSIMD as they train for each specific target identity.
Regarding Emotion Accuracy($\uparrow$), our approach has achieved superior performance compared to all existing methods on MEAD.

In contrast, our approach stands apart by requiring fine-tuning on only one video for unfamiliar identities, setting it apart from EVP and MEAD. 
Both of these methods need sample images depicting the target across various emotional states for training identity-specific models. 
This characteristic also serves as a distinct advantage of our model.

During evaluation on the CREMA-D dataset, despite the divergence in data sources, EmoTalker still attains the highest score in Emotion Accuracy, with only a marginal reduction in identity CSIMD.
\begin{table}[h]
\vspace{-1em} 
\centering
\caption{ Quantitative comparison of ours and the SOTA.}
\label{tab:metrics_comparison}
\begin{tabular}{l|l|cc}
\toprule
\multirow{2}{*}{dataset} & \multirow{2}{*}{model} & \multicolumn{1}{c}{Identity} & \multirow{2}{*}{EmoAcc($\uparrow$)} \\
& & \multicolumn{1}{c}{CSIMD($\downarrow$)} & \\

\midrule
\multirow{3}{*}{MEAD} & MEAD \cite{wang2020mead} & 0.86 & 76.00 \\
 & EVP \cite{Ji_2021_CVPR} & \textbf{0.67} & 83.58 \\
 & ours & 0.76 & \textbf{84.76} \\
\midrule
\multirow{3}{*}{CREMA-D} & Vougioukas \cite{vougioukas2020realistic} & \textbf{0.51} & 55.26 \\
 & Eskimez \cite{eskimez2021speech} & 0.75 & 65.67 \\
 & ours & 0.73 & \textbf{75.13} \\
\bottomrule
\end{tabular}
\vspace{-2em} 
\end{table}

\subsection{Expressions Generation via Prompts Containing Complex Emotions}
\begin{figure}[h]
    \centering
     \includegraphics[scale=0.325]{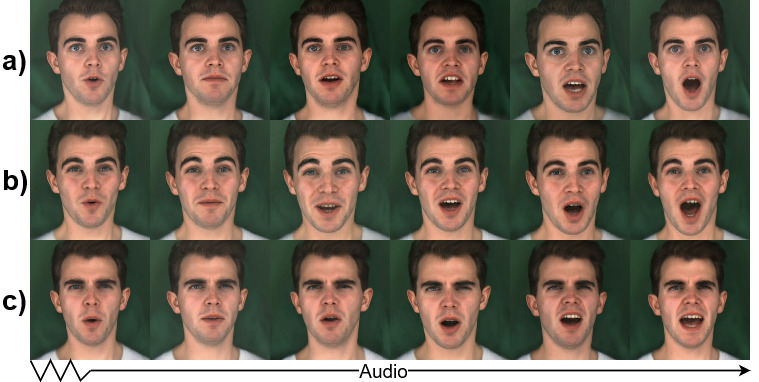}
    \caption{Expressions generation with different prompts.}
    \label{img1}
\end{figure}

The Figure~\ref{img1} illustrates the EmoTalker's performance in generating faces with the same neutral audio based on prompts that encompass intricate emotions.
The prompts associated with the expressions within the figure are: (a) fearful with a little sad, (b) very happy with a little surprised, and (c) very angry and disgusted.
It is evident that the generated facial expressions correspond to the intricate emotions encompassed in the given prompts. 

\vspace{-20pt} 
\subsection{Ablation Study}
\begin{figure}[t]
    \centering
     \includegraphics[scale=0.38]{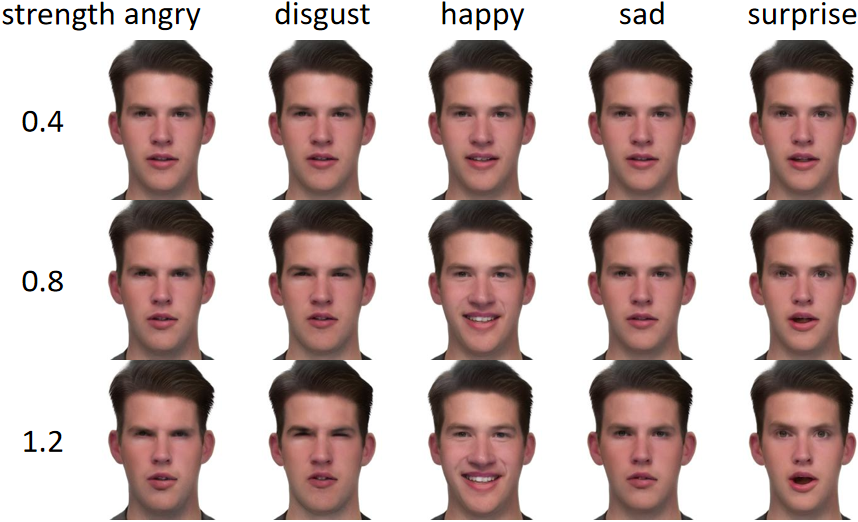}
    \caption{Expressions generation with different strengths. }
    \vspace{-15pt} 
    \label{img2}
\end{figure}

\textbf{Different Strength of Emotions.}
The Figure~\ref{img2} shows the performance of EmoTalker in generating faces based on different strengths driven by the same neutral audio.
It's apparent that different strengths of conditions can influence the degree of facial changes. 
With increasing strength, the alterations in facial expressions become more distinct.

\textbf{Denoising with Preserving Facial information.}
Table \ref{tab2} illustrates the influence of variations in the preservation ratio $a$ on identity CSIMD and Emotion Accuracy.
As the preserving ratio $a$ increases, the identity CSIMD decreases.
However, the robustness of generating accurate expressions decreases significantly.
To attain a substantial decrease in identity CSIMD without experiencing an obvious drop in Emotion Accuracy, we opt for a value of 0.3 for $a$ to preserve the facial information.

\begin{table}[h]
\vspace{-10pt} 
\centering
\caption{ Quantitative comparison of the preserving ratio a.}
\label{tab2}
\begin{tabular}{l|c|cc}
\toprule
dataset & ratio $a$ & Identity CSIMD($\downarrow$) & EmoAcc($\uparrow$) \\

\midrule
\multirow{6}{*}{MEAD} 
    & 0 & 0.87 & 86.32 \\
    & 0.1 & 0.83 & 85.69 \\
    & 0.2 & 0.79 & 85.22 \\
    & 0.3 & 0.76 & 84.76 \\
    & 0.4 & 0.70 & 81.13 \\
    & 0.5 & 0.68 & 76.35 \\
\bottomrule
\end{tabular}
\vspace{-2em} 
\end{table}

\section{Conclusion}
 In conclusion, this paper present EmoTalker, a diffusion-based emotionally editable talking face generation framework. 
 Through the introduction of preserving facial information during inference, we improve identity generalization capability of the model.
 Through the utilization of a curated dataset for emotion comprehension, the Emotion Intensity Block has learned complex emotional information to guide the generation of intricate expressions.
 As a results, EmoTalker achieves significant advancements in generating high-quality and customizable expressive talking faces. 

\vspace{-10pt} 
\section{Acknowledgement}
Supported by the Key Research and Development Program of Guangdong Province under grant No.2021B0101400003, the Natural Science Foundation of China (62276242), National Aviation Science Foundation (2022Z071078001), CAAI-Huawei MindSpore Open Fund (CAAIXSJLJJ-2022-001A), Anhui Province Key Research and Development Program (202104a05020007), Dreams Foundation of Jianghuai Advance Technology Center (2023-ZM01Z001). 
Corresponding authors are Ning Cheng from Ping An Technology (Shenzhen) Co., Ltd and Jun Yu from USTC.

\bibliographystyle{IEEEbib}
\bibliography{strings,refs}

\end{document}